# Robust Perception Architecture Design for Automotive Cyber-Physical Systems

Joydeep Dey, Sudeep Pasricha
Department of Electrical and Computer Engineering
Colorado State University, Fort Collins, CO, United States
{joydeep.dey, sudeep}@colostate.edu

## ABSTRACT

In emerging automotive cyber-physical systems (CPS), accurate environmental perception is critical to achieving safety and performance goals. Enabling robust perception for vehicles requires solving multiple complex problems related to sensor selection/ placement, object detection, and sensor fusion. Current methods address these problems in isolation, which leads to inefficient solutions. We present PASTA, a novel framework for global co-optimization of deep learning and sensing for dependable vehicle perception. Experimental results with the Audi-TT and BMW-Minicooper vehicles show how PASTA can find robust, vehicle-specific perception architecture solutions.

## 1. Introduction

In 2021, it was reported that an estimated 31,730 people died in motor vehicle traffic crashed in the United States, representing an estimated increase of about 12 percent compared to 2020 [1]. By eliminating the possibility of human driving errors through automation, advanced driver assistance systems (ADAS) are becoming a critical component in modern vehicles, to help save lives, improve fuel efficiency, and enhance driving comfort. ADAS systems typically involve a 4-stage pipeline involving sequential execution of functions related to *perception*, decision, control, and actuation. An incorrect understanding of the environment by the perception system can make the entire system prone to erroneous decision making, which can result in accidents due to imprecise real-time control and actuation. This motivates the need for a reliable *perception architecture* that can mitigate errors at the source of the pipeline and improve safety in emerging semi-autonomous vehicles.

The capabilities of a perception architecture for a vehicle depend on the SAE autonomy level (defined by the SAE-J3016 standard) supported by the vehicle. In general, an optimal vehicle perception architecture should consist of carefully defined location and orientation of each sensor selected from a heterogeneous suite of sensors (e.g., cameras, radars) to maximize environmental coverage. In addition to ensuring accurate sensing via appropriate sensor placement, a high object detection rate and low false positive detection rate needs to be maintained using efficient deep learning-based object detection and sensor fusion techniques.

State-of-the-art deep learning-based object detection models are built with different network architectures, uncertainty modeling approaches, and test datasets over a wide range of evaluation metrics [2]. For real-time perception, object detectors are resource-constrained by latency requirements, onboard memory capacity, and computational complexity. Optimizations performed to meet any one of these constraints often results in a trade-off with the performance of others [3]. As a result, selecting the best deep learning-based object detector for perception applications remains a challenge.

In real-world driving scenarios, the position of obstacles and traffic are highly dynamic, so after detection of an object, tracking is necessary to predict its new position. Due to noise from various sources, there is an inherent uncertainty associated with the measured position and velocity. This uncertainty is minimized by using sensor fusion algorithms [4]. A challenge with sensor fusion is that the complexity of tracking objects increases as the objects get closer, due to a much lower margin for error in the vicinity of the vehicle.

As summarized in Fig. 1, the design space of a vehicular perception architecture involves determining appropriate sensor selection and placement, object detection algorithms, and sensor fusion techniques. The possible configurations for each of these decisions is non-trivial and can easily lead to a combinatorial explosion of the design space, making exhaustive exploration impractical. Conversely, an optimization of each of these decisions individually before composing a final solution can lead to solutions that are sub-optimal and perform poorly in real environments. Today there are no generalized rules for the synthesis of perception architectures for automotive CPS, because perception architecture design depends heavily on the target features and use cases to be supported in the vehicle, which makes the already massive design space involved with the problem even larger and harder to traverse.

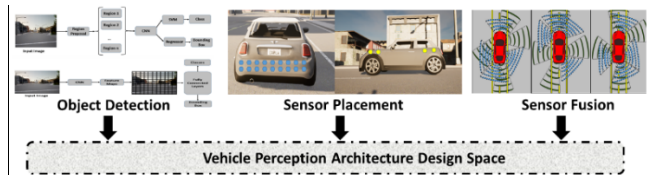

**Figure 1: Breakdown of perception architecture design space**

In this work, we propose a new framework called PASTA (Perception Architecture Search Technique for Automotive CPS) to perform perception architecture synthesis for semi-autonomous vehicles. *To the best of our knowledge, this is the first work to comprehensively explore and synthesize the sensing, fusion, and object detection perception subsystems jointly.* Our results indicate that PASTA can optimize perception performance across different vehicle types. The main contributions in this work include:

- A global co-optimization framework capable of synthesizing robust vehicle-specific perception architecture solutions that include heterogeneous sensor placement, deep learning-based object detector design, and sensor fusion algorithm selection;
- An exploration of various design space search algorithms tuned for the vehicle perception architecture search problem;
- A fast and efficient method for co-exploration of the deep learning object detector hyperparameters, through adaptive environment- and vehicle-specific transfer learning;
- A comparative analysis of the framework efficiency across different vehicle models (Audi TT, BMW Minicooper).

## 2. Related Work

State-of-the-art semi-autonomous vehicles require robust perception of their environment, for which the choice of sensor placement, object detection algorithms, and sensor fusion techniques are the most important decisions. These decisions are carefully curated to support ADAS features (e.g., blindspot warning, lane keep assist) that characterize the autonomy level to be supported.

Many prior works have explored vehicle perception system design with different combinations of sensor types to overcome limitations that plague individual sensor types. For example, [5] used a single camera-radar pair for perception of headway distance using a radar mounted on the geometric center of the front bumper and a monocular camera behind the windscreen. In [6] a camera-radar based perception architecture was proposed for target acquisition with the well-known SSD (Single Shot Detection) object detector on consecutive camera frames. The detection accuracy was optimized with the use of a Kalman filter and Bayesian estimation, which reduced computational complexity compared to [5]. In [7] a single neural network (CRF-Net) was proposed for detection and fusion of all camera and radar detections. The work in [8] optimized merging camera detection with LiDAR data. A fusion scheme was used to sequentially merge 2D detections made by a YOLOv3 object detector using cylindrical projection with the detections made from LiDAR point cloud data. In [9], an approach to fuse LiDAR and stereo camera data was proposed. In contrast to the projection of 2D detections in [8], [9] used a projection of 3D LiDAR points into the camera image frame instead, which upsampled the projection image, creating a more dense depth map. *All these prior works optimize vehicle perception for rigid combinations of sensors and object detectors, without any design space exploration.*

Only a few prior works have (partially) explored the design space of sensors and object detectors for vehicle perception. An approach for optimal positioning and calibration of a three LiDAR system was proposed in [10]. The approach used a neural network to learn and qualify the effectiveness of different LiDAR location and orientations. In [11] a sensor selection and exploration approach was proposed based on factor graphs during multi-sensor fusion. The work in [12] heuristically explored a subset of backbone networks in the Faster R-CNN object detector for perception systems in vehicles. [13] presented a framework that used a genetic algorithm to optimize sensor orientations and placements in vehicles. *Unlike prior works that fine-tune specific perception architectures, e.g., [5]-[9], or explore the sensing and object detector configurations separately, e.g. [10]-[13], this paper proposes a holistic framework that jointly co-optimizes heterogeneous sensor placement, object detection algorithms, and sensor fusion techniques.* To the best of our knowledge, this is the first work that performs co-optimization across such a comprehensive decision space to optimize vehicle perception, with support for deployment across multiple vehicle types.

## 3. Background
### 3.1 ADAS Level 2 Autonomy Features

In this work, our exploration of perception architectures on a vehicle, henceforth referred to as an *ego vehicle*, targets four ADAS features that have varying degrees of longitudinal (i.e., in the same lane as the ego vehicle) and lateral (i.e., in neighboring lanes to the ego vehicle lane) sensing requirements. The SAE-J3016 standard [14] defines adaptive cruise control (ACC) and lane keep assist (LKA) individually as level 1 features, as they only perform the dynamic driving task in either the latitudinal or longitudinal direction of the vehicle. Forward collision warning (FCW) and blindspot warning (BW) are defined in SAE-J3016 as level 0 active safety systems, as they only enhance the performance of the driver without performing any portion of the dynamic driving task. However, when all four features are combined, the system can be described as a level 2 autonomy system. Fig. 2 shows an overview of the four features we focus on for level 2 autonomy, which are discussed next.

Although implementations differ, all **ACC** (adaptive cruise control) systems take over longitudinal control from the driver (Fig. 2). The challenge in ACC is to maintain an accurate track of the lead vehicle (immediately ahead of the ego vehicle in the same lane) with a forward-facing sensor and using longitudinal control to maintain the specified distance while maintaining driver comfort (e.g., avoiding sudden velocity changes). **LKA** (lane keep assist) systems determine whether the ego vehicle is drifting towards any lane boundaries. LKA systems have been known to over-compensate, creating a "ping-pong" effect where the vehicle oscillates back and forth between the lane lines [15]. The main challenges in LKA are to reduce this ping-pong effect and the accurate detection of lane lines on obscured (e.g., snow covered) roads. **FCW** (forward collision warning) systems are used for real-time prediction of collisions with a lead vehicle. It is important that this system avoids false positives as well as false negatives to improve driver comfort, safety, and reduce rear end accidents [16]. Lastly, **BW** (blindspot warning) systems use lateral sensor data to determine whether there is a vehicle towards the rear on either side of the ego vehicle (Fig. 2) in a location the driver cannot see with their side mirrors. *A perception architecture designed to support Level 2 autonomy in a vehicle should support all four of these critical features.*

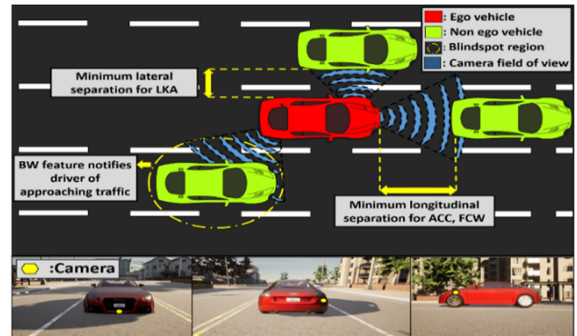

**Figure 2: Common scenarios in ACC, FCW, LKA, and BW**

### 3.2 Sensor Placement and Orientation

For capturing the most relevant data pertaining to each feature, sensors need to be placed strategically on the ego vehicle to maximize coverage of the vehicle environment needed for a feature. Fig. 2 shows an example of field of view coverage (in blue) corresponding to three unique placements of camera sensors on the body of the ego vehicle (in yellow, lower images) to meet coverage goals. For the ACC and FCW features, the ego vehicle is responsible for slowing down to maintain a minimum separation between the ego and lead vehicle. The camera must be positioned somewhere on the front bumper to measure minimum longitudinal separation accurately while keeping the lead vehicle in the desired field of view. For LKA, there is a need to maintain a safe minimum lateral distance between non-ego vehicles in neighboring lanes. Here a front camera is needed to extract lane line information, while side cameras are required for tracking this minimum lateral separation. BW requires information about a specific area near the rear of the vehicle, requiring sensor placement that maximizes the view of the blind spot. If the sensor is too far forward or too far back, it will miss key portions of the blind spots. Beyond placement, the orientation of sensors can also significantly impact coverage for all features [13]. *Thus, sensor placement and orientation remains a challenging problem.*

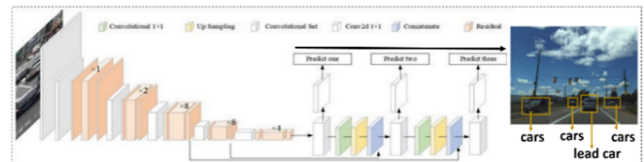

**Figure 3: Example of vehicle (object) detection with YOLOv3**

### 3.3 Object Detection for Vehicle Environment Perception

There are two broad goals associated with deep learning-based object detectors: 1) determining spatial information (relative position of an object in the image) via *localization*, and 2) identifying which category that object instance belongs to via *classification* [17]. Fig. 3 shows object detection of multiple car instances using the YOLOv3 object detector [18] by creating a bounding box around the 'car' object instances and predicting the object class as 'car'. The pipeline of object detection models can be divided into informative region selection, feature extraction, and classification [19]. Depending on which subset of these steps are used to process an input image frame, object detectors are classified as single-stage or two-stage.

Modern <u>single-stage</u> detectors are composed of a fully convolutional neural network (CNN) that outputs object classification probabilities and box offsets (w.r.t. pre-defined anchor/ bounding boxes) at each spatial position. The YOLO family of object detectors is a popular example of single-stage detectors [18]. SSD (single shot detection) is another example, based on the VGG-16 backbone [20]. An advantageous property of single-stage detectors is their very high detection throughput (e.g., ~40 frames per second with YOLO) that makes them suitable for real time scenarios. <u>Two-stage</u> detectors divide the detection process into separate region proposal and classification stages. In the first stage, several regions in an image that have a high probability to contain an object are identified with a region proposal network (RPN). In the second stage, proposals of identified regions are fed into CNNs for classification. Region-based CNN (R-CNN) is an example of a two-stage detector [21]. R-CNN divides an input image into 2000 regions generated through a selective search algorithm, after which the selected regions are fed to a CNN for feature extraction followed by classification. Fast R-CNN [22] and subsequently Faster R-CNN [23] improved the speed of training as well as detection accuracy compared to R-CNN by streamlining the stages. Two-stage detectors have high localization and object recognition accuracy, whereas one-stage detectors achieve higher inference speed [24]. *In our work, we consider both types of object detectors to exploit the latency/accuracy tradeoffs during perception architecture synthesis.*

### 3.4 Sensor Fusion

Perception architectures that use multiple sensors must deal with errors due to imprecise measurements from one or more of the sensors. Conversely, errors can also arise when only a single sensor is used due to measurement uncertainties from insufficient spatial (*occlusion*) or temporal (*delayed sensor response time*) coverage of the environment. The Kalman filter is one of the most widely used sensor fusion state estimation algorithms that enables error-resilient tracking of targets [25]. Kalman filters have the ability to obtain optimal statistical estimations when the system state is described as a linear model and the error can be modeled as Gaussian noise. If the system state is represented as a nonlinear dynamic model as opposed to a linear model, a modified version of the Kalman filter known as the Extended Kalman Filter (EKF) can be used, which provides an optimal approach for implementing nonlinear recursive filters [26]. However, the computation of the Jacobian (matrix describing the system state) in EKF can be computationally expensive [27]. The unscented Kalman filter (UKF) is another alternative that has the property of being more amenable to parallel implementation [28]. *In our perception architecture exploration, we explore the family of Kalman filters as candidates for sensor fusion.*

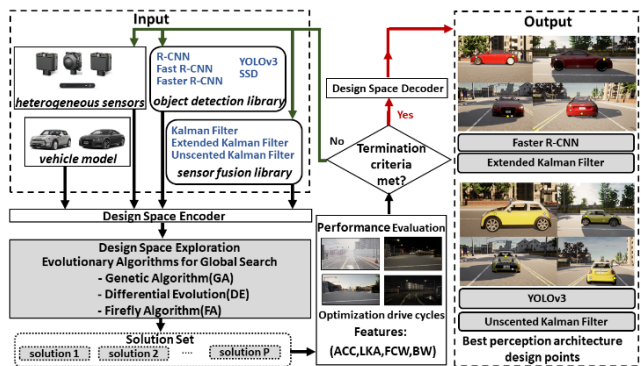

**Figure 4: An overview of the proposed PASTA framework**

## 4. PASTA Framework
### 4.1 Overview

Fig. 4 presents a high-level overview of our proposed *PASTA* framework. The heterogeneous sensors, object detection model library, sensor fusion algorithm library, and physical dimensions of the vehicle model are inputs to the framework. An algorithmic design space exploration is used to generate a perception architecture solution which is subsequently evaluated based on a cumulative score from performance metrics relevant to the ADAS autonomy level being targeted. We evaluate three design space search exploration algorithms as part of the framework: genetic algorithm (GA), differential evolution (DE), and the firefly algortham (FA). The process of perception architecture generation and evaluation iterates until an algorithm-specific stopping criteria is met, at which point the best design points are output. The following subsections describe each component of our framework in detail.

### 4.2 Problem Formulation and Metrics

In our framework, for a given vehicle, we define a design point as a perception architecture that is a combination of three components: a sensor configuration (with a deployment position and orientation of each sensor selected for the vehicle), an object detector algorithm, and a sensor fusion algorithm. *The goal is to find an optimal design point for the given vehicle that minimizes the cumulative error across eight metrics that are characteristic of the ability to track and detect non-ego vehicles across road geometries and traffic scenarios.* The eight selected metrics are related to a goal of supporting level 2 autonomy with the perception architecture (note that PASTA can be easily extended to higher autonomy levels by adding additional metrics and features). In the descriptions of the metrics below, the ground truth refers to the actual position of the non-ego vehicles (traffic in the environment of the ego vehicle).

The metrics can be summarized as: *1) longitudinal position error and 2) lateral position error:* deviation of the detected positional data from the ground truth of non-ego vehicle positions along the y and x axes, respectively; *3) object occlusion rate:* the fraction of passing non-ego vehicles that go undetected in the vicinity of the ego vehicle; *4) velocity uncertainty:* the fraction of times that the velocity of a non-ego vehicle is measured incorrectly; *5) rate of late detection:* the fraction of the number of 'late' non-ego vehicle detections made over the total number of non-ego vehicles. Late detection is one that occurs after a non-ego vehicle crosses the minimum safe longitudinal or lateral distance, as defined by safety models for pre-crash scenarios [29]. This metric directly factors in the trade-off between latency and accuracy for object detector and fusion algorithms; *6) false positive lane detection rate:* the fraction of instances when a lane marker is detected but there exists no ground truth lane; *7) false negative lane detection rate:* the fraction of instances when a ground truth lane exists but is not detected; and *8) false positive object detection rate:* the fraction of total vehicle detections which were classified as non-ego vehicle detections but did not actually exist.

### 4.3 Design Space Encoder/Decoder

The design space encoder receives a set of random initial design points which are encoded into a vector format, best suited for various kinds of rearrangement and splitting operations during design space exploration. The encoder adapts the initial selection of inputs for our design space such that a design point is defined by the location and orientation of each sensor's configuration (consisting of six parameters: x, y, z, roll, pitch, and yaw), together with the object detector and fusion algorithm. The design space decoder converts the solutions into the same format as the input so that the output perception architecture solution(s) found can be visualized with respect to the real-world co-ordinate system.

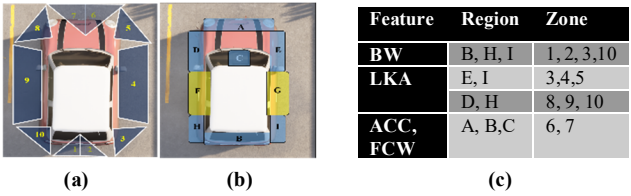

| Feature | Region | Zone |
|---|---|---|
| BW | B, H, I | 1, 2, 3, 10 |
| LKA | E, I | 3, 4, 5 |
| | D, H | 8, 9, 10 |
| ACC, FCW | A, B, C | 6, 7 |

(a)            (b)            (c)

**Figure 5:** (a) Field of view (FOV) zones; (b) sensor placement regions; (c) feature, region, and zone relationship

### 4.4 Design Space Exploration

The goal of a design space exploration algorithm in our framework is to generate perception architectures (design points) which are aware of feature to field of view (FOV) zone correlations around an ego vehicle. Fig. 5(a) shows the 10 primary FOV zones around the ego-vehicle. These zones of interest are defined as the most important perception areas in the environment for a particular ADAS feature. Fig. 5(b) shows the regions on the vehicle on which sensors can be mounted (in blue). Regions F and G (in yellow) are exempt from sensor placement due to the mechanical instability of placing sensors on the door of a vehicle. The correlation between ADAS features, zones, and regions, is shown in Fig. 5(c). For exploration of possible locations within a region, a fixed step size of 2cm in two dimensions across the surface of the vehicle is considered, which generates a 2D grid of possible positions in each zone shown in Fig. 5(b). The orientation exploration of each sensor involves rotation at a fixed step size of 1 degree between an upper and lower bounding limit for roll, pitch, and yaw respectively, at each of these possible positions within the 2D grid. The orientation exploration limits were chosen with caution with the caveat that some sensors, such as long-range radars, have an elevated number of recorded false positives with extreme orientations.

To get a sense of the large design space, consider four sensors (e.g., 2 cameras, 2 radars). Just the determination of the optimal placement and orientation of these sensors involves exploring $^{1.24e+26}C_4$ and $^{7.34e+25}C_4$ configurations for the Audi-TT and BMW-Minicooper vehicles, respectively. Coupled with the choice of different object detectors and sensor fusion algorithms, *the resulting massive design space cannot be exhaustively traversed in a practical amount of time, necessitating the use of intelligent design space search algorithms that support hill climbing to escape local minima.*

In our framework, we explored three evolutionary algorithms: 1) *Genetic Algorithm (GA)*, 2) *Differential Evolution (DE)*, and 3) *Firefly Algorithm (FA)*. As shown in Fig. 4, each algorithm generates a solution set of size *'P'* at every iteration until the termination criteria is met. The algorithms simultaneously co-optimize sensor configuration, object detection, and sensor fusion, and also explore new regions of the design space when the termination (perception) criteria is not met. The algorithms are briefly described below.

*4.1.1 Genetic Algorithm (GA)*

GA is a popular evolutionary algorithm to solve optimization problems by mimicking the process of natural selection [30]. GA repeatedly selects a population of candidate solutions and then improves the solutions by modifying them. GA can optimize problems where the design space is discontinuous and also if the cost function is non-differentiable. In our GA implementation, in the selection stage, the cost function values are computed for 50 design points at a time, and a roulette wheel selection method is used to select which set of chromosomes will be involved in the crossover step based on their cost function probability value (fraction of the cumulative cost function sum of all chromosomes considered in the selection). In the crossover stage, the crossover parameter is set to 0.5, allowing half of the 50 chromosomes to produce offspring. The mutation parameter is set to 0.2 which determines the new genes allowed for mutation in each iteration.

*4.1.2 Differential Evolution (DE)*

Differential Evolution (DE) [31] is another stochastic population-based evolutionary algorithm that takes a unique approach to mutation and recombination. An initial solution population of fixed size is selected randomly, and each solution undergoes mutation and then recombination operations. DE generates new parameter vectors by adding the weighted difference between two population vectors to a third vector to achieve difference vector-based mutation. Next, crossover is performed, where the mutated vector's parameters are mixed with the parameters of another vector (called the target vector), to yield a trial vector. If the trial vector yields a lower cost function value than the target vector, the trial vector replaces the target vector in the next generation. Greedy selection is performed between the target vector and trial vector at every iteration to ensure better solutions are selected only after generation of all trial vectors. Unlike GA where parents are selected based on fitness, every solution in DE takes turns to be one of the parents [32]. In our DE implementation, we set initial population size to 50 and use a crossover probability of 0.8 to select candidates participating in crossover.

*4.1.3 Firefly Algorithm (FA)*

FA is a swarm-based metaheuristic [33] that has shown superior performance compared to GA for certain problems [34]. In FA, a solution is referred to as a firefly. The algorithm mimics how fireflies interact using flashing lights (bioluminescence). The algorithm assumes that all fireflies can be attracted by any other firefly. Further, the attractiveness of a firefly is directly proportional to its brightness which depends on the fitness function value. Initially, a random solution set is generated and the fitness (brightness) of each candidate solution is measured. In the design space, a firefly is attracted to another with higher brightness (more fit solution), with brightness decreasing exponentially over distance. FA is significantly different from DE and GA, as both exploration of new solutions and exploitation of existing solutions to find better solutions is achieved using a single position update step.

### 4.5 Performance Evaluation

Each design point in the solution set generated per iteration of the design space exploration undergoes performance evaluation across drive cycles. A *drive cycle* here refers to a virtual simulation involving an ego-vehicle (with a perception architecture under evaluation) following a fixed set of waypoint co-ordinates, while performing object detection and sensor fusion on the environment and other non-ego vehicles. A total of 20 different drive cycles were considered, with 5 drive cycles customized for each ADAS feature. As an example, drive cycles for ACC and FCW involve an ego vehicle following different lead vehicles at different distances, velocities, weather conditions, and traffic profiles. The fitness of the perception architectures generated by PASTA are computed using the cumulative metric scores (Section 4.2) across the drive cycles.

## 5. Experiments
### 5.1 Experimental Setup

To evaluate the efficacy of the PASTA framework we performed experiments in the open-source simulator CARLA implemented as a layer on Unreal Engine 4 (UE4) [35]. The UE4 engine provides state-of-the-art physics rendering for highly realistic driving scenarios. We used this tool to design multiple *optimization drive cycles* that are roughly 5 minutes long using maps that included linear and curved highway lanes, T junctions, rural environments, Michigan lefts in urban highways and 4-way traffic intersections. The ego vehicle follows a fixed route for ~3 miles in each map and encounters scenarios commonly found in real driving environments, such as adverse weather conditions (rain, fog) and overtly aggressive/conservative driving styles observed with some vehicles. To ensure generalizability in evaluating solution quality we consider a separate set of *test drive cycles* where the ego vehicle travels approximately 3 miles in 5 minutes. The maps and ego vehicle trajectories selected for test drive cycles are different from the optimization drive cycles. We target generating perception architectures to meet level 2 autonomy goals for two vehicles: Audi-TT and BMW-Minicooper (Fig. 6). PASTA can be extended for higher autonomy levels by adding additional metrics and features and can also easily target other vehicles if provided with their critical dimensions, as in Fig. 6.

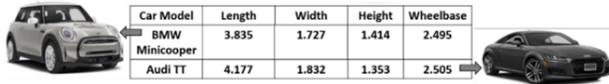

Figure 6: BMW Minicooper (left) and Audi TT (right)

The design space considered for evaluation uses a maximum of 8 sensors: up to 4 radars and 4 cameras that can be placed in any zone (Fig. 5(a)-(b)). Using a greater number of these sensors led to negligible improvements for the level 2 autonomy goal. The RGB cameras possess 90º field of view, 200 fps shutter speed, and image resolution of 800×600 pixels. The mid-range radars selected generate a maximum of 1500 measurements per second with a horizontal and vertical field of view of 30º and a maximum detection distance of 100 meters. We considered 5 different object detectors (YOLOv3, SSD, R-CNN, Fast R-CNN, and Faster R-CNN) and 3 sensor fusion algorithms (Kalman filter, Extended Kalman filter, and Unscented Kalman filter). For the design space exploration algorithms, the cost function was a weighted sum across the eight metrics (Section 4.2), with the weight factor for each metric chosen based on their total feature-wise cardinality across all zones shown in Fig. 5(c). During design space exploration, if the change in average cost function value was < 5% over 250 iterations, the search was terminated. All algorithmic exploration was performed on an AMD Ryzen 7 3800X 8-Core CPU desktop with an NVIDIA GeForce RTX 2080 Ti GPU.

## 5.2 Experimental Results

In the first experiment, we explored the inference latency and accuracy in terms of mean average precision (mAP) for the five different object detectors considered. Table 1 shows the inference latency on a CPU and GPU, as well as the accuracy in mAP for the object detectors on images from our analyzed drive cycles, with all detectors trained on the MS-COCO dataset. It can be observed that the two-stage detectors (R-CNN, Fast R-CNN, Faster R-CNN) have a higher accuracy than the single stage detectors (SSD, YOLOv3). However, the inference time for two-stage detectors is significantly higher than for the single stage detectors. For real-time object detection in vehicles, it is crucial to be able to detect objects with low latency, typically less than 100ms [37]. As a result, single stage detectors are preferable, with YOLOv3 achieving slightly better accuracy and lower inference time than SSD. However, in some scenarios, delayed detection can still be better than not detecting or wrongly detecting an object (e.g., slightly late blindspot warning is still better than receiving no warning) in which case the slower but more accurate two-stage detectors may still be preferable. In PASTA, we therefore explore both single-stage and two-stage detectors, and factor in the accuracy and rate of late detection in our metrics (Section 4.2). Also, detectors with a higher mAP value sometimes did not detect objects that other detectors with a lower mAP were able to; thus we consider all five detectors in our exploration.

Table 1: Object detector latency and accuracy comparison

| Obj. Detector | Latency GPU (ms) | Latency CPU (ms) | mAP(%) |
|---|---|---|---|
| R-CNN | 48956.18 | 66090.83 | 73.86 |
| Fast R-CNN | 1834.71 | 2365.86 | 76.81 |
| Faster R-CNN | 176.99 | 286.72 | 79.63 |
| SSD | 53.25 | 70.32 | 70.58 |
| YOLOv3 | 24.03 | 32.92 | 71.86 |

Next, we explored the importance of global co-optimization for our problem. We select the genetic algorithm (GA) variant of our framework to explore the entire design space (GA-PASTA) and compared it against five other frameworks. Frameworks GA-PO and GA-OP use the GA but perform a local (sequential) search for sensor design. In GA-PO, sensor position is explored before orientation, while in GA-OP the orientation for fixed sensor locations (based on industry best practices) is explored before adjusting sensor positions. For both frameworks, the object detector used was fixed to YOLOv3 due to its sub-100ms inference latency and reasonable accuracy, while the extended Kalman filter (EKF) was used for sensor fusion due to its ability to efficiently track targets following linear or non-linear trajectories. The framework GA-VESPA is from prior work [13] and uses GA for exploring sensor positions and orientations simultaneously, with the YOLOv3 object detector and EKF fusion algorithm. Frameworks GA-POD and GA-POF use GA for a more comprehensive exploration of the design space. GA-POD simultaneously explores the sensor positioning, orientation, and object detectors, with a fixed EKF fusion algorithm. GA-POF simultaneously explores the sensor positioning, orientation, and sensor fusion algorithm, with a fixed YOLOv3 fusion algorithm.

Fig. 7(a) depicts the average cost of solution populations (lower is better) for the BMW-Minicooper across the different frameworks plotted against the number of iterations, with each exploration lasting between 80-100 hours. It can be observed that GA-PO outperforms GA-OP, which confirms the intuitive importance of exploring sensor positioning before adjusting sensor orientations. GA-VESPA outperforms both GA-PO and GA-OP, highlighting the benefit of co-exploration of sensor position and orientation over a local sequential search approach used in GA-PO and GA-OP. GA-POD and GA-POF in turn outperform these frameworks, indicating that decisions related to object detection and sensor fusion can have a notable impact on perception quality. GA-POD terminates with its solution set having a lower average cost than GA-POF, which indicates that co-exploration of object detection and sensor placement/orientation is slightly more effective than co-exploration of sensor fusion and sensor placement/orientation. *Our GA-PASTA framework achieves the lowest average cost solution, highlighting the tremendous benefit that can be achieved from co-exploring sensor position/orientation, object detection, and sensor fusion algorithms.*

Fig. 7(b) summarizes the objective function cost of the best solution found by each framework, which aligns with the population-level observations from Fig. 7(a). The comparative analysis for the BMW-Minicooper was repeated three times with different initializations for all six frameworks, and the results for the other two runs show a consistent trend with the one shown in Fig. 7. Note also that the relative trend across frameworks observed for the Audi-TT is similar to that observed for the BMW-Minicooper, and thus the results for the Audi TT are omitted for brevity.

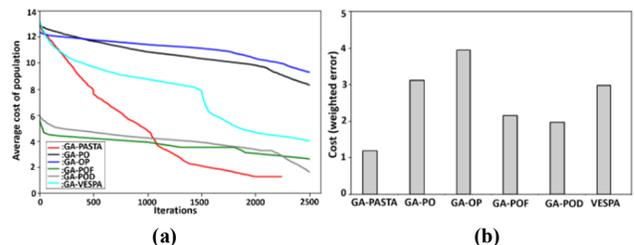

Figure 7: (a) Comparison of perception architecture exploration frameworks; (b) Cost of best solution from each framework

In the next experiment, we explored the effectiveness of different design space exploration algorithms (GA, DE, and FA; see Section 4.4) to determine which algorithm can provide optimal perception architecture solutions across varying vehicle models. Fig. 8 shows the results for the three variants of the PASTA framework, for the Audi-TT and BMW-Minicooper vehicles. The best solution was selected across three runs of each algorithmic variant (variations for the best solution across runs are highlighted with confidence intervals, with bars indicating the median). The FA algorithm outperforms the DE and GA algorithms for both vehicles. For Audi-TT, the best solution found by FA improves upon the best solution found with DE and GA by 18.34% and 14.84%, respectively. For the BMW-Minicooper the best solution found by FA outperforms the best solution found by DE and GA by 3.16% and 13.08%, respectively. Fig. 9(a) depicts the specific sensor placement locations for each vehicle type, with a visualization of sensor coverage for the best solutions found by each algorithm shown in fig. 9(b).

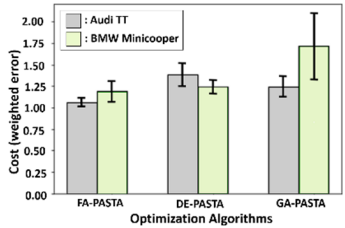

**Figure 8:** Comparison of three variants of PASTA with genetic algorithm (GA), differential evolution (DE), and Firefly algorithm (FA)

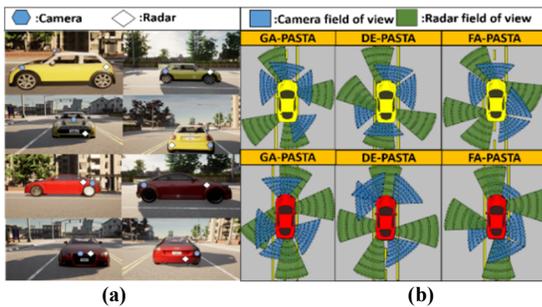

**Figure 9:** (a) Sensor placement for best solution found with FA algorithm (top yellow vehicle: BMW-Minicooper, bottom red vehicle: Audi-TT); (b) Sensor coverage for best solutions found by GA, DE, and FA algorithms

Lastly, in our quest to further improve perception architecture synthesis in PASTA, we focused on improved exploration of the object detector design space. We selected the FA algorithm due to its superior performance over GA and DE, and modified FA-PASTA to integrate a neural architecture search (NAS) for the YOLOv3 object detector, to further improve YOLOv3 accuracy while maintaining its low detection latency. Our NAS for YOLOv3 involved transfer learning to retrain network layers with a dataset consisting of 6000 images obtained from the KITTI dataset, using the open-source tool CADET [36]. The NAS hyperparameters that were explored involved the number of layers to unfreeze and retrain (from a total of 53 layers in the Darknet-53 backbone used in YOLOv3; Fig. 10(a)), along with the optimizer learning rate, momentum, and decay. The updated variant of our framework, FA-NAS-PASTA, considered these YOLOv3 hyperparameters along with the sensor positions and orientations, and sensor fusion algorithms, during iterative evolution of the population of candidate solutions in the FA algorithm.

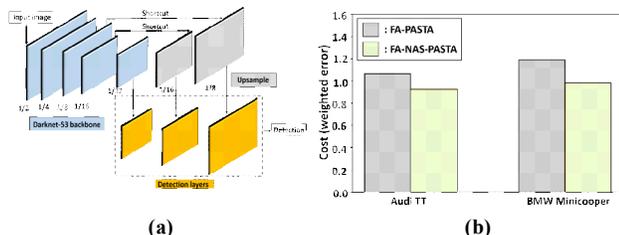

**Figure 10:** (a) YOLOv3 object detector architecture with Darknet-53 backbone network that was fine-tuned using neural architecture search (NAS); (b) results of integrating object detector NAS with PASTA

Fig. 10(b) shows the results of this analysis. FA-PASTA is the best performing variant of our framework (from Fig. 8), while FA-NAS-PASTA is the modified variant that integrates NAS for YOLOv3. It can be observed that fine tuning the YOLOv3 object detector during search space exploration in FA-NAS-PASTA leads to notable improvements in the best perception architecture solution, with up to 14.43% and 21.13% improvement in performance for the Audi-TT and BMW-Minicooper, compared to PASTA-FA.

## 6. Conclusions

In this work, we propose a new framework called *PASTA* that can generate perception architectures for semi-autonomous automotive cyber-physical systems. *PASTA* can simultaneously co-optimize locations and orientations for sensors, object detectors, and sensor fusion algorithms for any given vehicle. Our analysis showed how PASTA can synthesize optimized perception architectures for the Audi TT and BMW Minicooper vehicles, while outperforming multiple semi-global exploration techniques. Integrating neural architecture search (NAS) for the object detector in PASTA shows further promising improvements in solution quality.